\documentclass[11pt,a4paper]{article}
\usepackage[hyperref]{acl2019}
\usepackage{times}
\usepackage{latexsym}
\usepackage{url}
\usepackage{pst-node}
\usepackage{mathtools}
\usepackage{amssymb, amsmath}
\usepackage{tikz}
\usetikzlibrary{topaths}
\usepackage{array, multirow, boldline}
\usepackage{mathtools}

\usepackage[caption=false]{subfig}
\usepackage{hyperref}
\DeclarePairedDelimiter\abs{\lvert}{\rvert}
\newlength{\savecw}

\aclfinalcopy 
\def\aclpaperid{104} 

\title{Reducing Gender Bias in Word-Level Language Models with a Gender-Equalizing  Loss Function}

\author{Yusu Qian\thanks{Yusu Qian and Urwa Muaz contributed equally to the paper.} 
\\
  Tandon School \\of Engineering \\ New York University\\
  6 MetroTech Center\\Brooklyn, NY, 11201 \\
  \texttt{yq729@nyu.edu} \\\And
  Urwa Muaz\footnotemark[1] \\
  Tandon School \\of Engineering \\ New York University\\
  6 MetroTech Center\\Brooklyn, NY, 11201\\
  \texttt{um367@nyu.edu}\And
  Ben Zhang \\
  Center for \\Data Science \\ New York University\\
  60 Fifth Avenue\\ New York, NY, 10012\\ 
  \texttt{bz957@nyu.edu} \\\And
  Jae Won Hyun\\
  Department of \\Computer Science\\
  New York University\\
  251 Mercer St\\ New York, NY, 10012\\
  \texttt{jaewhyun@nyu.edu}}

\date{}
\tikzstyle{every picture}+=[remember picture,inner xsep=0,inner ysep=0.25ex]

\begin{document}
\maketitle
\begin{abstract}
Gender bias exists in natural language datasets which neural language models tend to learn, resulting in biased text generation. In this research, we propose a debiasing approach based on the loss function modification. We introduce a new term to the loss function which attempts to equalize the probabilities of male and female words in the output. Using an array of bias evaluation metrics, we provide empirical evidence that our approach successfully mitigates gender bias in language models without increasing perplexity by much. In comparison to existing debiasing strategies, data augmentation, and word embedding debiasing, our method performs better in several aspects, especially in reducing gender bias in occupation words. Finally, we introduce a combination of data augmentation and our approach, and show that it outperforms existing strategies in all bias evaluation metrics.
\end{abstract}

\section{Introduction}

Natural Language Processing (NLP) models are shown to capture unwanted biases and stereotypes found in the training data which raise concerns about socioeconomic, ethnic and gender discrimination when these models are deployed for public use \citep{luetal, zhaoetal}. 

There are numerous studies that identify algorithmic bias in NLP applications. \citet{vilesuggestions} showed ethnic bias in Google autocomplete suggestions whereas \citet{lambrecht} found gender bias in advertisement delivery systems. Additionally, \citet{zhaoetal} demonstrated that coreference resolution systems exhibit gender bias.

Language modelling is a pivotal task in NLP with important downstream applications such as text generation \citep{sutskever}. Recent studies by \citet{luetal} and \citet{bordia} have shown that this task is vulnerable to gender bias in the training corpus. Two prior works focused on reducing bias in language modelling by data preprocessing \citep{luetal} and word embedding debiasing \citep{bordia}. In this study, we investigate the efficacy of bias reduction during training by introducing a new loss function which encourages the language model to equalize the probabilities of predicting gendered word pairs like \textit{he} and \textit{she}. Although we recognize that gender is non-binary, for the purpose of this study, we focus on female and male words.

Our main contributions are summarized as follows: i) to our best knowledge, this study is the first one to investigate bias alleviation in text generation by direct modification of the loss function; ii) our new loss function effectively reduces gender bias in the language models during training by equalizing the probabilities of male and female words in the output; iii) we show that end-to-end debiasing of the language model can achieve word embedding debiasing; iv) we provide an interpretation of our results and draw a comparison to other existing debiasing methods. We show that our method, combined with an existing method, counterfactual data augmentation, achieves the best result and outperforms all existing methods.

\section{Related Work}

Recently, the study of bias in NLP applications has received increasing attention from researchers. Most relevant work in this domain can be broadly divided into two categories: word embedding debiasing and data debiasing by preprocessing.

\paragraph{Word Embedding Debiasing} \citet{bolukbasi} introduced the idea of gender subspace as low dimensional space in an embedding that captures the gender information. \citet{bolukbasi} and \citet{zhao2017} defined gender bias as a projection of gender-neutral words on a gender subspace and removed bias by minimizing this projection. \citet{gonen} proved that bias removal techniques based on minimizing projection onto the gender space are insufficient. They showed that male and female stereotyped words cluster together even after such debiasing treatments. Thus, gender bias still remains in the embeddings and is easily recoverable. 

\citet{bordia} introduced a co-occurrence based metric to measure gender bias in texts and showed that the standard datasets used for language model training exhibit strong gender bias. They also showed that the models trained on these datasets amplify bias measured on the model-generated texts. Using the same definition of embedding gender bias as \citet{bolukbasi}, \citet{bordia} introduced a regularization term that aims to minimize the projection of neutral words onto the gender subspace. Throughout this paper,we refer to this approach as REG. They found that REG reduces bias in the generated texts for some regularization coefficient values. But, this bias definition is shown to be incomplete by \citet{gonen}. Instead of explicit geometric debiasing of the word embedding, we implement a loss function that minimizes bias in the output and thus adjust the whole network accordingly. For each model, we analyze the generated word embedding to understand how it is affected by output debiasing.

\paragraph{Data Debiasing}  \citet{luetal} showed that gender bias in coreference resolution and language modelling can be mitigated through a data augmentation technique that expands the corpus by swapping the gender pairs like \textit{he} and \textit{she}, or \textit{father} and \textit{mother}. They called this Counterfactual Data Augmentation (CDA) and concluded that it outperforms the word embedding debiasing strategy proposed by \citet{bolukbasi}. CDA doubles the size of the training data and increases time needed to train language models. In this study, we intend to reduce bias during training without requiring an additional data preprocessing step. 

\section{Methodology}

\subsection{Dataset}
For the training data, we use Daily Mail news articles released by \citet{hermann}. This dataset is composed of 219,506 articles covering a diverse range of topics including business, sports, travel, etc., and is claimed to be biased and sensational \citep{bordia}.  For manageability, we randomly subsample 5\% of the text. The subsample has around 8.25 million tokens in total.

\subsection{Language Model}
We use a pre-trained 300-dimensional word embedding, GloVe, by \citet{pennington}. We apply random search to the hyperparameter tuning of the LSTM language model. The best hyperparameters are as follows: 2 hidden layers each with 300 units, a sequence length of 35, a learning rate of 20 with an annealing schedule of decay starting from 0.25 to 0.95, a dropout rate of 0.25 and a gradient clip of 0.25. We train our models for 150 epochs, use a batch size of 48, and set early stopping with a patience of 5. 
\subsection{Loss Function}
Language models are usually trained using cross-entropy loss. Cross-entropy loss at time step \begin{math}t\end{math} is
\begin{equation*}
L^{CE}(t) = -\sum_{w\in{V}}y_{w,t}\log{(\hat{y}_{w,t})}\,,
\end{equation*}
\noindent where \begin{math}V\end{math} is the vocabulary, \begin{math}y\end{math} is the one hot vector of ground truth and \begin{math}\hat y\end{math} indicates the output softmax probability of the model.  

We introduce a loss term \begin{math}L^B\end{math}, which aims to equalize the predicted probabilities of gender pairs such as \textit{woman} and \textit{man}. 

\begin{equation*}
L^B(t) =\frac{1}{G}\sum_i^G\abs*{\log{\frac{\hat{y}_{f_i,t}}{\hat{y}_{m_i,t}}}}    
\end{equation*}
\begin{math}f\end{math} and \begin{math}m\end{math} are a set of corresponding gender pairs, $G$ is the size of the gender pairs set, and \begin{math}\hat{y}\end{math} indicates the output softmax probability. We use gender pairs provided by \citet{zhao2017}. By considering only gender pairs we ensure that only gender information is neutralized and distribution over semantic concepts is not altered. For example, it will try to equalize the probabilities of \textit{congressman} with \textit{congresswoman} and \textit{actor} with \textit{actress} but distribution of {\textit{congressman}, \textit{congresswoman}} versus {\textit{actor}, \textit{actress}} will not be affected. Overall loss can be written as
\begin{equation*}
L=\frac{1}{T}\sum_{t=1}^TL^{CE}(t) + \lambda{}L^B(t)\,,    
\end{equation*}
where \begin{math}\lambda\end{math} is a hyperparameter and $T$ is the corpus size. We observe that among the similar minima of the loss function, \begin{math}L^B\end{math} encourages the model to converge towards a minimum that exhibits the lowest gender bias.

\subsection{Model Evaluation}
Language models are evaluated using perplexity, which is a standard measure of performance for unseen data. For bias evaluation, we use an array of metrics to provide a holistic diagnosis of the model behavior under debiasing treatment. These metrics are discussed in detail below. In all the evaluation metrics requiring gender pairs, we use gender pairs provided by \citet{zhao2017}. This list contains 223 pairs, all other words are considered gender-neutral.

\begin{table*}[ht]
    \centering
    \setlength{\savecw}{\columnwidth}
    \subfloat[Occupation bias conditioned on gendered words \label{tab:firstset}]{
    \begin{minipage}{\savecw}
    \centering
            \begin{tabular}{c c c c}
            \multirow{1}{*}{} &
            \multirow{1}{*}{\tikz[baseline=(node1.base)]\node (node1)  {\textbf{He}}; is \tikz[baseline=(node2.base)]\node (node2) {a}; $|$} &
            \multirow{3}{*}{ \tikz[baseline=(node5.base)]\node (node5)  {\textbf{d}};\textbf{octo}\tikz[baseline=(node6.base)]\node (node6) {\textbf{r}};} & \multirow{3}{*}{$\log{\frac{P(t|s_1)}{P(t|s_2)}}$} \\
            \multirow{1}{*}{} \\ & 
            \multirow{1}{*}{\tikz[baseline=(node3.base)]\node (node3)  {\textbf{She}}; is \tikz[baseline=(node4.base)]\node (node4) {a}; $|$} & &\\
         \end{tabular}
    \begin{tikzpicture}[remember picture, overlay]
        \draw [bend right] (node2.north) to node [above] {$s_1$} (node1.north);
        \draw [bend right] (node4.north) to node [above] {$s_2$} (node3.north);
        \draw [bend right] (node6.north) to node [above] {$t$} (node5.north);
    \end{tikzpicture}
    \end{minipage}
    }
    \hfill
    \subfloat[Occupation bias conditioned on occupations\label{tab:secondset}]{
    \begin{minipage}{\savecw}
    \centering
    \begin{tabular}{c c c c}
            \multirow{3}{*}{} & 
            \multirow{3}{*}{\tikz[baseline=(node7.base)]\node(node7) {The}; \textbf{doctor} is \tikz[baseline=(node8.base)]\node(node8) {a}; $|$} &
            \multirow{1}{*}{ \tikz[baseline=(node9.base)]\node (node9) {\textbf{m}};\textbf{a}\tikz[baseline=(node10.base)]\node (node10) {\textbf{n}};} & \multirow{3}{*}{$\log{\frac{P(t_1|s)}{P(t_2|s)}}$} \\ \\
             & & \multirow{1}{*}{ \tikz[baseline=(node11.base)]\node (node11) {\textbf{w}};\textbf{oma}\tikz[baseline=(node12.base)]\node (node12) {\textbf{n}};} &
    \end{tabular}
    \begin{tikzpicture}[remember picture, overlay]
        \draw [bend right] (node8.north) to node [above] {$s$} (node7.north);
        \draw [bend right] (node10.north) to node [above] {$t_1$} (node9.north);
        \draw [bend right] (node12.north) to node [above] {$t_2$} (node11.north);
    \end{tikzpicture}
    \end{minipage}}
    \caption{Example templates of two types of occupation bias\label{tab:overallset}}
\end{table*}

\subsubsection{Co-occurrence Bias}
Co-occurrence bias is computed from the model-generated texts by comparing the occurrences of all gender-neutral words with female and male words. A word is considered to be biased towards a certain gender if it occurs more frequently with words of that gender. This definition was first used by \citet{zhao2017} and later adapted by \citet{bordia}. Using the definition of gender bias similar to the one used by \citet{bordia}, we define gender bias as
\begin{equation*}
    B^N = \frac{1}{N}\sum_{w\in N}\abs*{\log\frac{c(w,m)}{c(w,f)}}\,,
\end{equation*}
where \begin{math}N\end{math} is a set of gender-neutral words, and \begin{math}c(w,g)\end{math} is the occurrences of a word \begin{math}w\end{math} with words of gender \begin{math}g\end{math} in the same window. This score is designed to capture unequal co-occurrences of neutral words with male and female words. Co-occurrences are computed using a sliding window of size 10 extending equally in both directions. Furthermore, we only consider words that occur more than 20 times with gendered words to exclude random effects. 

We also evaluate a normalized version of \begin{math}B^N\end{math} which we denote by conditional co-occurrence bias, \begin{math}B_c^N\end{math}. This is defined as

\begin{equation*}
    B_c^N = \frac{1}{N}\sum_{w\in N}\abs*{\log\frac{P(w|m)}{P(w|f)}}\,,
\end{equation*}
where
\begin{equation*}
    P(w|g)=\frac{c(w,g)}{c(g)}\,.
\end{equation*}
\begin{math}B_c^N\end{math} is less affected by the disparity in the general distribution of male and female words in the text. The disparity between the occurrences of the two genders means that text is more inclined to mention one over the other, so it can also be considered a form of bias.
We report the ratio of occurrence of male and female words in the model generated text, \begin{math}GR\end{math}, as 
\begin{equation*}
    GR=\frac{c(m)}{c(f)}\,.
\end{equation*}

\subsubsection{Causal Bias}
Another way of quantifying bias in NLP models is based on the idea of causal testing. The model is exposed to paired samples which differ only in one attribute (e.g. gender) and the disparity in the output is interpreted as bias related to that attribute. \citet{zhaoetal} and \citet{luetal} applied this method to measure bias in coreference resolution and \citet{luetal} also used it for evaluating gender bias in language modelling. 

Following the approach similar to \citet{luetal}, we limit this bias evaluation to a set of gender-neutral occupations. We create a list of sentences based on a set of templates. There are two sets of templates used for evaluating causal occupation bias (Table \ref{tab:overallset}). The first set of templates is designed to measure how the probabilities of occupation words depend on the gender information in the seed. Below is an example of the first set of templates:
\begin{equation*}
    [Gendered\,word]\,is\,a\,|\,[occupation]\,.
\end{equation*}
\noindent Here, the vertical bar separates the seed sequence that is fed into the language models from the target occupation, for which we observe the output softmax probability. We measure causal occupation bias conditioned on gender as
\begin{equation*}
CB|g = \frac{1}{|O|}\frac{1}{G}\sum_{o\in O}\sum_i^G\abs*{\log\frac{p(o|f_i)}{p(o|m_i)}}\,,    
\end{equation*}
\noindent where \begin{math}O\end{math} is a set of gender-neutral occupations and \begin{math}G\end{math} is the size of the gender pairs set. For example, \begin{math}P(doctor|he)\end{math} is the softmax probability of the word \begin{math}doctor\end{math} where the seed sequence is \textit{He is a}. 
The second set of templates like below, aims to capture how the probabilities of gendered words depend on the occupation words in the seed.
\begin{equation*}
    The\,[occupation]\,is\,a\,|\,[gendered\,word]\,.
\end{equation*}
Causal occupation bias conditioned on occupation is represented as 
\begin{equation*}
    CB|o = \frac{1}{|O|}\frac{1}{G}\sum_{o\in O}\sum_i^G\abs*{\log\frac{p(f_i|o)}{p(m_i|o)}}\,,
\end{equation*}
\noindent where \begin{math}O\end{math} is a set of gender-neutral occupations and \begin{math}G\end{math} is the size of the gender pairs set. For example, \begin{math}P(man|doctor)\end{math} is the softmax probability of \textit{man} where the seed sequence is \textit{The doctor is a}. 

We believe that both \begin{math}CB|g\end{math} and \begin{math}CB|o\end{math} contribute to gender bias in the model-generated texts. We also note that \begin{math}CB|o\end{math} is more easily influenced by the general disparity in male and female word probabilities.

\subsubsection{Word Embedding Bias}
Our debiasing approach does not explicitly address the bias in the embedding layer. Therefore, we use gender-neutral occupations to measure the embedding bias to observe if debiasing the output layer also decreases the bias in the embedding. We define the embedding bias, \begin{math}EB_d\end{math}, as the difference between the Euclidean distance of an occupation word to male words and the distance of the occupation word to the female counterparts. This definition is equivalent to bias by projection  described by \citet{bolukbasi}. We define \begin{math}EB_d\end{math} as
\begin{equation*}
\begin{split}
    EB_d=\sum_{o\in O}\sum_i^G|\|E(o)-E(m_i)\|_2
    \\-\|E(o)-E(f_i)\|_2|\,,
\end{split}
\end{equation*}
\noindent where \begin{math}O\end{math} is a set of gender-neutral occupations, \begin{math}G\end{math} is the size of the gender pairs set and \begin{math}E\end{math} is the word-to-vector dictionary.

\subsection{Existing Approaches}
We apply CDA where we swap all the gendered words using a bidirectional dictionary of gender pairs described by \citet{luetal}. This creates a dataset twice the size of the original data, with exactly the same contextual distributions for both genders and we use it to train the language models. 

We also implement the bias regularization method of \citet{bordia} which debiases the word embedding during language model training by minimizing the projection of neutral words on the gender axis. We use hyperparameter tuning to find the best regularization coefficient and report results from the model trained with this coefficient. We later refer to this strategy as REG.

\begin{table*}[h]
    \centering
    \begin{tabular}{c c c c c c c c c}
    \hline\hline\\[-2.5ex]
    Model & \begin{math}B^N\end{math} & \begin{math}B_c^N\end{math} & \begin{math}GR\end{math} & \begin{math}Ppl.\end{math} & \begin{math}CB|o\end{math} & \begin{math}CB|g\end{math} &\begin{math}EB_d\end{math} \\
    \hline
    Dataset & 0.340 & 0.213 &  &  - & - & - & - \\
    Baseline & 0.531 & 0.282 & 1.415 &  117.845 & 1.447 & 97.762 & 0.528 \\
    REG & 0.381 & 0.329 & 1.028 & \textbf{114.438} & 1.861 & 108.740 & 0.373 \\
    CDA & 0.208 & 0.149 & 1.037 & 117.976 & 0.703 & 56.82 & 0.268 \\
    \begin{math}\lambda_{0.01}\end{math} & 0.492  & 0.245 & 1.445 & 118.585 & 0.111 & 9.306 & 0.077 \\
    \begin{math}\lambda_{0.1}\end{math} & 0.459  & 0.208 & 1.463 & 118.713 & 0.013 & 2.326 & 0.018 \\
    \begin{math}\lambda_{0.5}\end{math} & 0.312  & 0.173 & 1.252 & 120.344 & \textbf{0.000} & 1.159 & 0.006 \\
    \begin{math}\lambda_{0.8}\end{math} & 0.226  & 0.151 & 1.096 & 119.792 & 0.001 & 1.448 & 0.002 \\
    \begin{math}\lambda_{1}\end{math} & 0.218  & 0.153 & 1.049 & 120.973 & \textbf{0.000} & 0.999 & 0.002 \\
    \begin{math}\lambda_{2}\end{math} & 0.221  & 0.157 & 1.020 & 123.248 & \textbf{0.000} & 0.471 & \textbf{0.000} \\
    $\boldsymbol\lambda_{\textbf{0.5}}$ \textbf{+ CDA} & \textbf{0.205} & \textbf{0.145} & \textbf{1.012} & 117.971 & \textbf{0.000} & \textbf{0.153} & \textbf{0.000} \\ 
    \end{tabular}
    \caption{Evaluation results for models trained on Daily Mail and their generated texts}
    \label{tab:table2}
\end{table*}

\section{Experiments}
Initially, we measure the co-occurrence bias in the training data. After training the baseline model, we implement our loss function and tune for the \begin{math}\lambda\end{math} hyperparameter. We test the existing debiasing approaches, CDA and REG, as well but since \citet{bordia} reported that results fluctuate substantially with different REG regularization coefficients, we perform hyperparameter tuning and report the best results in Table \ref{tab:table2}. Additionally, we implement a combination of our loss function and CDA and tune for \begin{math}\lambda\end{math}. Finally, bias evaluation is performed for all the trained models. Causal occupation bias is measured directly from the models using template datasets discussed above and co-occurrence bias is measured from the model-generated texts, which consist of 10,000 documents of 500 words each.

\subsection{Results}
Results for the experiments are listed in Table \ref{tab:table2}. It is interesting to observe that the baseline model amplifies the bias in the training data set as measured by \begin{math}B^N\end{math}and \begin{math}B_c^N\end{math}. From measurements using the described bias metrics, our method effectively mitigates bias in language modelling without a significant increase in perplexity. At \begin{math}\lambda\end{math} value of 1, it reduces \begin{math}B^N\end{math} by 58.95\%, \begin{math}B_c^N\end{math} by 45.74\%, \begin{math}CB|o\end{math} by 100\%, \begin{math}CB|g\end{math} by 98.52\% and \begin{math}EB_d\end{math} by 98.98\%. Compared to the results of CDA and REG, it achieves the best results in both occupation biases, \begin{math}CB|g\end{math} and \begin{math}CB|o\end{math}, and \begin{math}EB_d\end{math}. We notice that all methods result in \begin{math}GR\end{math} around 1, indicating that there are near equal amounts of female and male words in the generated texts. In our experiments we note that with increasing \begin{math}\lambda\end{math}, the bias steadily decreases and perplexity tends to slightly increase. This indicates that there is a trade-off between bias and perplexity.

REG is not very effective in mitigating bias when compared to other methods, and fails to achieve the best result in any of the bias metrics that we used. But REG results in the best perplexity and even does better than the baseline model in this respect. This indicates that REG has a slight regularization effect. Additionally, it is interesting to note that our loss function outperforms REG in \begin{math}EB_d\end{math}  even though REG explicitly aims to reduce gender bias in the embeddings. Although our method does not explicitly attempt geometric debiasing of the word embedding, the results show that it results in the most debiased embedding as compared to other methods. Furthermore, \citet{gonen} emphasizes that geometric gender bias in word embeddings is not completely understood and existing word embedding debiasing strategies are insufficient. Our approach provides an appealing end-to-end solution for model debiasing without relying on any measure of bias in the word embedding.  We believe this concept is generalizable to other NLP applications. 

Our method outperforms CDA in \begin{math}CB|g\end{math}, \begin{math}CB|o\end{math}, and \begin{math}EB_d\end{math}. While CDA achieves slightly better results for co-occurrence biases, \begin{math}B^N\end{math}and \begin{math}B_c^N\end{math}, and results in a better perplexity. With a marginal differences,  our results are comparable to those of CDA and both models seem to have similar bias mitigation effects. However, our method does not require a data augmentation step and allows training of an unbiased model directly from biased datasets. For this reason, it also requires less time to train than CDA since its training data has a smaller size without data augmentation. Furthermore, CDA fails to effectively mitigate occupation bias when compared to our approach. Although the training data for CDA does not contain gender bias, the model still exhibits some gender bias when measured with our causal occupation bias metrics. This reinforces the concept that some model-level constraints are essential to debiasing a model and dataset debiasing alone cannot be trusted.

Finally, we note that the combination of CDA and our loss function outperforms all the methods in all measures of biases without compromising perplexity. Therefore, it can be argued that a cascade of these approaches can be used to optimally debias the language models.

\section{Conclusion and Discussion}
In this research, we propose a new approach for mitigating gender bias in neural language models and empirically show its effectiveness in reducing bias as measured with different evaluation metrics. Our research also highlights the fact that debiasing the model with bias penalties in the loss function is an effective method. We emphasize that loss function based debiasing is powerful and generalizable to other downstream NLP applications. The research also reinforces the idea that geometric debiasing of the word embedding is not a complete solution for debiasing the downstream applications but encourages end-to-end approaches to debiasing. 

All the debiasing techniques experimented in this paper rely on a predefined set of gender pairs in some way. CDA used gender pairs for flipping, REG uses it for  gender space definition and our technique uses them for computing loss. This reliance on pre-defined set of gender pairs can be considered a limitation of these methods. It also results in another concern. There are gender associated words which do not have pairs, like pregnant. These words are not treated properly by techniques relying on gender pairs.

Future work includes designing a context-aware version of our loss function which can distinguish between the unbiased and biased mentions of the gendered words and only penalize the biased version. Another interesting direction is exploring the application of this method in mitigating racial bias which brings more challenges. 

\section{Acknowledgment}
We are grateful to Sam Bowman for helpful advice, Shikha Bordia, Cuiying Yang, Gang Qian, Xiyu Miao, Qianyi Fan, Tian Liu, and Stanislav Sobolevsky for discussions, and reviewers for detailed feedback.
\bibliographystyle{acl_natbib}
\nocite{*}
\bibliography{acl2019.bib}

\begin{thebibliography}{11}
\expandafter\ifx\csname natexlab\endcsname\relax\def\natexlab#1{#1}\fi

\bibitem[{Bolukbasi et~al.(2016)Bolukbasi, Chang, Zou, Saligrama, and
  Kalai}]{bolukbasi}
Tolga Bolukbasi, Kai-Wei Chang, James Zou, Venkatesh Saligrama, and Adam Kalai.
  2016.
\newblock \href {https://arxiv.org/pdf/1607.06520.pdf} {Man is to computer
  programmer as woman is to homemaker? debiasing word embeddings}.
\newblock In \emph{NIPS'16 Proceedings of the 30th International Conference on
  Neural Information Processing Systems}, pages 4356--4364.

\bibitem[{Bordia and Bowman(2019)}]{bordia}
Shikha Bordia and Samuel~R. Bowman. 2019.
\newblock \href {https://arxiv.org/pdf/1904.03035.pdf} {Identifying and
  reducing gender bias in word-level language models}.
\newblock ArXiv:1904.03035.

\bibitem[{Gonen and Goldberg(2019)}]{gonen}
Hila Gonen and Yoav Goldberg. 2019.
\newblock \href {https://arxiv.org/pdf/1903.03862.pdf} {Lipstick on a pig:
  Debiasing methods cover up systematic gender biases in word embeddings but do
  not remove them}.
\newblock ArXiv:1903.03862.

\bibitem[{Hermann et~al.(2015)Hermann, Kočiský, Grefenstette, Espeholt, Kay,
  Suleyman, and Blunsom}]{hermann}
Karl Hermann, Tomáš Kočiský, Edward Grefenstette, Lasse Espeholt, Will Kay,
  Mustafa Suleyman, and Phil Blunsom. 2015.
\newblock \href {https://arxiv.org/pdf/1506.03340.pdf} {Teaching machines to
  read and comprehend}.
\newblock In \emph{NIPS'15 Proceedings of the 28th International Conference on
  Neural Information Processing Systems}, pages 1693--1701.

\bibitem[{Lambrecht and Tucker(2018)}]{lambrecht}
Anja Lambrecht and Catherine~E. Tucker. 2018.
\newblock \href {https://ssrn.com/abstract=2852260} {Algorithmic bias? an
  empirical study into apparent gender-based discrimination in the display of
  stem career ads}.

\bibitem[{Lapowsky(2018)}]{vilesuggestions}
Issie Lapowsky. 2018.
\newblock \href
  {https://www.wired.com/story/google-autocomplete-vile-suggestions/} {Google
  autocomplete still makes vile suggestions}.

\bibitem[{Lu et~al.(2018)Lu, Mardziel, Wu, Amancharla, and Datta}]{luetal}
Kaiji Lu, Piotr Mardziel, Fangjing Wu, Preetam Amancharla, and Anupam Datta.
  2018.
\newblock \href {https://arxiv.org/pdf/1807.11714.pdf} {Gender bias in neural
  natural language processing}.
\newblock ArXiv:1807.11714v1.

\bibitem[{Pennington et~al.(2014)Pennington, Socher, and Manning}]{pennington}
Jeffrey Pennington, Richard Socher, and Christopher Manning. 2014.
\newblock \href {https://doi.org/10.3115/v1/D14-1162} {Glove: Global vectors
  for word representation}.
\newblock In \emph{Proceedings of the 2014 Conference on Empirical Methods in
  Natural Language Processing}, page 1532–1543. Association for Computational
  Linguistics.

\bibitem[{Sutskever et~al.(2011)Sutskever, Martens, and Hinton}]{sutskever}
Ilya Sutskever, James Martens, and Geoffrey Hinton. 2011.
\newblock \href {https://dl.acm.org/citation.cfm?id=3104610} {Generating text
  with recurrent neural networks}.
\newblock In \emph{ICML'11 Proceedings of the 28th International Conference on
  International Conference on Machine Learning}, pages 1017--1024.

\bibitem[{Zhao et~al.(2017)Zhao, Wang, Yatskar, Ordonez, and Chag}]{zhao2017}
Jieyu Zhao, Tianlu Wang, Mark Yatskar, Vicente Ordonez, and Kai-Wei Chag. 2017.
\newblock \href {https://arxiv.org/pdf/1707.09457.pdf} {Men also like shopping:
  Reducing gender bias amplification using corpus-level constraints}.
\newblock In \emph{Conference on Empirical Methods in Natural Language
  Processing}.

\bibitem[{Zhao et~al.(2018)Zhao, Zhou, Li, Wang, and Kaiwei}]{zhaoetal}
Jieyu Zhao, Yichao Zhou, Zeyu Li, Wei Wang, and Chang Kaiwei. 2018.
\newblock \href {https://www.aclweb.org/anthology/D18-1521} {Learning
  gender-neutral word embeddings}.
\newblock In \emph{Proceedings of the 2018 Conference on Empirical Methods in
  Natural Language Processing}, page 4847–4853. Association for Computational
  Linguistics.

\end{thebibliography}
\end{document}